\documentclass{article}

\usepackage[final]{nips_2017}

\usepackage[utf8]{inputenc} 
\usepackage[T1]{fontenc}    
\usepackage{hyperref}       
\usepackage{url}            
\usepackage{booktabs}       
\usepackage{amsfonts}       
\usepackage{amsmath}       
\usepackage{nicefrac}       
\usepackage{microtype}      
\usepackage{graphicx}

\graphicspath{ {images/} }

\setcitestyle{square}

\title{Visualizing Information Bottleneck through Variational Inference}

\author{
  Cipta Herwana\\
  New York University\\
  \texttt{ciptah@nyu.edu}\\
  \And
  Abhishek Kadian\\
  New York University\\
  \texttt{abhishekk@nyu.edu} \\
}

\begin{document}

\maketitle

\begin{abstract}
  The Information Bottleneck theory provides a theoretical and computational framework for finding approximate minimum sufficient statistics. Analysis of the Stochastic Gradient Descent (SGD) training of a neural network on a toy problem has shown the existence of two phases, fitting and compression. In this work, we analyze the SGD training process of a Deep Neural Network on MNIST classification and confirm the existence of two phases of SGD training. We also propose a setup for estimating the mutual information for a Deep Neural Network through Variational Inference.
\end{abstract}

\section{Introduction}\label{section:intro}

Deep Neural Networks (DNNs) have found wide application on large-scale tasks like visual object recognition (\citet{DBLP:journals/cacm/KrizhevskySH17}), machine translation (\citet{DBLP:journals/corr/WuSCLNMKCGMKSJL16}) and reinforcement learning (\citet{DBLP:journals/nature/SilverHMGSDSAPL16}). The success of DNNs in many areas has led to a growing interest in trying to explain their performance. \citet{DBLP:journals/corr/TishbyZ15} proposed to analyze DNNs through the Information Bottleneck lens. \citet{DBLP:journals/corr/Shwartz-ZivT17} analyzed the information plane of a small neural network on a toy problem and reported two phases of a neural network trained using SGD, fitting and compression. In our work we test the hypothesis of Information Bottleneck theory for Deep Learning on a tougher problem of image classification.

For our experiments we need a way to estimate the mutual information between the input and the output of a hidden layer. \citet{DBLP:journals/corr/AlemiFD016} proposed a variational approximation to the Information Bottleneck. We also build upon the Variational Autoencoder derivation (\citet{kingma2013autovae}) to propose an alternative MI upper bound for a teacher-student model.

We use the variational estimates to test the Information Bottleneck hypothesis of two phases of DNN training, fitting and compression. We find that the two phases are present in the training of a 4 layer DNN on MNIST classification task. Our results show that the claims of Information Bottleneck theory about Deep Learning hold true for a deep VIB network.

Our paper begins with a discussion of Information Bottleneck theory and it's application to analyze DNNs (section-\ref{section:ib-and-dl}). In section-\ref{section:problem-definition} we define the classification problem and leveraging variational inference to calculate mutual information. We also derive the mutual information upper bounds in section-\ref{section:problem-definition}. In section-\ref{section:experiments} we discuss the experimental settings and report results on the classification task. We deliver our concluding remarks in section-\ref{section:conclusion} and also suggest future research directions.

\section{Information Bottleneck and Deep Learning}\label{section:ib-and-dl}

\subsection{Mutual Information}

Given any two random variables, $X$ and $Y$, with a joint distribution $p(x, y)$, their \emph{Mutual Information} is defined as:

\begin{equation}
  I(X;Y) = D_{KL}[p(x, y) || p(x) p(y)] = \sum \limits_{x \in X, y \in Y} p(x, y) \log \left(\frac{p(x, y)}{p(x) p(y)} \right)
\end{equation}

\begin{equation}
= \sum \limits_{x \in X, y \in Y} p(x, y) \log \left( \frac{p(x|y)}{p(x)} \right) = H(X) - H(X|Y)
\end{equation}

where $D_{KL}[p || q]$ is the Kullback-Liebler divergence of the distributions $p$ and $q$, and $H(X)$ and $H(X|Y)$ are the entropy and conditional entropy of $X$ and $Y$, respectively.

\subsection{Information Bottleneck}

When analyzing representations of $X$ w.r.t $Y$, the classical notion of \emph{minimal sufficient statistics} provide good candidates for optimal representation. Sufficient statistics, in our context, are maps or partitions of $X$, $S(X)$, that capture all the information that $X$ has on $Y$. Namely, $I(S(X); Y) = I(X; Y)$.

Minimal sufficient statistics, $T(X)$, are the simplest sufficient statistics and induce the coarsest sufficient partition on $X$. A simple of way of formulating this is through the Markov chain: $Y \rightarrow X \rightarrow S(X) \rightarrow T(X)$, which should hold for a minimal sufficient statistics $T(X)$ with any other sufficient statistics $S(X)$. Since exact minimal sufficient statistics only exist for very special distributions, (i.e. exponential families), \citet{tishby2000information} relaxed this optimization problem by first allowing the map to be stochastic, defined as an encoder $P(T|X)$, and then, by allowing the map to capture as much as possible of $I(X; Y)$, not necessarily all of it.

This leads to the \emph{Information Bottleneck} (IB) tradeoff [\citet{tishby2000information}], which provides a computational framework for finding approximate minimal sufficient statistics, or the optimal tradeoff between compression of $X$ and prediction of $Y$.

\subsection{Information Plane of Neural Nets}

Any representation variable, $T$, defined as a (possibly stochastic) map of the input $X$, is characterized by its joint distributions with $X$ and $Y$, or by its encoder and decoder distributions, $P(T|X)$ and $P(Y|T)$, respectively. Given $P(X; Y)$, $T$ is uniquely mapped to a point in the \emph{Information Plane} with coordinates $I(X; T), I(T; Y)$.

\citet{DBLP:journals/corr/TishbyZ15} proposed to analyze DNNs in the Information Plane and suggested that the goal of the neural network is to optimize the Information Bottleneck tradeoff between compression and prediction, successively, for each layer.

\begin{figure}
  \centering
    \includegraphics[width=150mm]{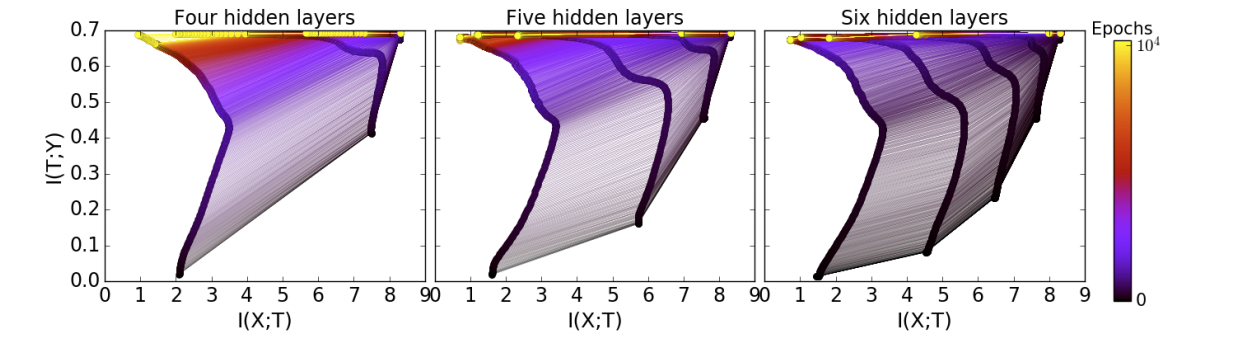}
    \caption{Results taken from \citet{DBLP:journals/corr/Shwartz-ZivT17}. The layers information paths during the SGD optimization for different architectures. Each panel is the information plane for a network with a different number of hidden layers. The width of the hidden layers start with 12, and each additional layer has 2 fewer neurons. The final layer with 2 neurons is shown in all panels. The line colors correspond to the number of training epochs.}
\label{figure:1}
\end{figure}

Building on top of this work \citet{DBLP:journals/corr/Shwartz-ZivT17} analyzed the information plane of a network with 7 fully connected hidden layers, and widths 12-10-7-5-4-3-2 neurons with hyperbolic tangent activations. The analysis showed that the Stochastic Gradient Descent (SGD) optimization has two main phases, in the first and shorter phase the layers increase the information on the input (fitting), while in the second much longer phase the layers reduce the information on the input (compression phase). The tasks were chosen as binary decision rules which are invariant under $O(3)$ rotations of the sphere, with 12 binary inputs that represent 12 uniformly distributed points on the sphere. As the network size was small they calculated the mutual information exhaustively by binning the output activations into 30 buckets and computing the joint distribution. Figure \ref{figure:1} shows the information plane for the described setup.

\citet{anonymous2018onrebut} showed that results of \citet{DBLP:journals/corr/Shwartz-ZivT17} are not applicable for networks with ReLU activations and also claim that the shape of information plane is a result of saturating hyperbolic tangent non-linearities and for ReLU activation the same shape is not observed. \citet{anonymous2018onrebut} follows a binning strategy to estimate the mutual information for ReLU activations. This method of estimating mutual information has been refuted by authors of \citet{DBLP:journals/corr/Shwartz-ZivT17}, ICLR-2018 Openreview\footnote{Review comments and discussion available at https://goo.gl/U24Kfp}.

In section 4 we discuss how we estimate mutual information using variational inference. Using these results we draw the information plane for a neural network learned to classify on the MNIST dataset\footnote{http://yann.lecun.com/exdb/mnist}. We use ReLU non-linearities and a deeper neural network for our task. In our results we observe the two distinct phases of SGD, fitting and compression as originally observed by \citet{DBLP:journals/corr/Shwartz-ZivT17}.

\section{Problem Definition}\label{section:problem-definition}

Information plane analysis is usually limited to toy problems with simple distributions, otherwise the MI calculation quickly becomes intractable. We examine approaches to estimating the information plane position using variational inference.

\subsection{Task: MNIST Classification}

\begin{figure}
  \centering
    \includegraphics[width=100mm]{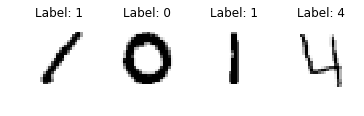}
  \caption{MNIST classification}
\label{figure:2}
\end{figure}

The MNIST classification task consists of images of digits and the task is to predict the label of digit. An example of the task is shown in in Figure \ref{figure:2}. We consider $X$ to be the input space (images) and $Y$ to be the output label space (digit labels). We modify this task into the teacher-student setting, where the digit images are generated by a pre-trained teacher model. 

\subsection{Variational Information Bottleneck Model}

The IB setup views the data generating process as a Markov chain $Y \rightarrow X \rightarrow Z \rightarrow \hat{Y}$. $Y$ is a signal we wish to extract from observations $X$. $Z$ is some intermediate representation computed from $Z$ by the model on the way to computing the predictions $\hat{Y}$.

The VIB \citep{DBLP:journals/corr/AlemiFD016} method is a generalization of the Variational Autoencoder \citep{kingma2013autovae} into the supervised setting. Its objective function is to maximize the rate distortion tradeoff:
$$
\mathcal{L}(\theta) = I(Y; Z) - \beta I(X; Z)
$$
These two quantities correspond to the position in the information plane, so computing an approximation of the loss will give us an estimate of the network's position. The MI of $Y$ (target variable) and $Z$ can be computed as the difference between entropy and conditional entropy (the remaining uncertainty after observing $Z$):
\begin{align*}
I(Y; Z) & = H(Y) - H(Y|Z) \\
& = H(Y) + \sum_{y, z} p(y, z) \log p(y|z)
\end{align*}
Under our setup $p(y|z)$ is difficult to compute, because we would need to invoke Bayes' rule on $p(z|y)$, which is what we have (after marginalizing over all $x$). So we introduce an approximate distribution $q(y|z)$:
\begin{align*}
I(Y; Z) & = H(Y) + \sum_{y, z} p(y, z) \log p(y|z) \frac{q(y|z)}{q(y|z)} \\
& = H(Y) + \sum_{y, z} p(y, z) \log q(y|z)
+ \sum_{y, z} p(y, z) \log \frac{p(y|z)}{q(y|z)} \\
& = H(Y) + \mathbb{E}_{y, z} \log q(y|z)
+ D_{KL}(p(y|z)||q(y|z)) \\
& \ge H(Y) + \mathbb{E}_{y, z} \log q(y|z)
\end{align*}

The second term is the average log-likelihood of $y$ under $q$. In practice we will use the learned decoder of the VIB model as the approximate $q$. The second term can be thought of as a conditional cross entropy term, which is an upper bound on the conditional entropy. To calculate $I(X; Z)$ we decompose it the same way:
\begin{align*}
I(X; Z) & = H(Z) - H(Z|X) \\
& = H(Z) -\mathbb{E}_{x}H[p(z|x)]
\end{align*}
$\mathbb{E}_{x}H[p(z|x)]$ is the average entropy of the conditional distribution $p(z|x)$ for a fixed $x$. Intuitively, this makes sense. If the entropy of the encoding is high, then there is high uncertainty of choosing $z$ from $x$, so mutual information should be low. Because $H(Z)$ is hard to compute, we substitute it with the cross-entropy of $p(z)$ and some variational approximation $r(z)$. Cross entropy is always greater than entropy, so:
\begin{align*}
I(X; Z) & \le - \sum_{x, z} p(x, z) \log r(z) + \sum_{x, z} p(x, z) \log p(z|x) \\
& = \sum_{x, z} p(x, z) \log \frac{p(z|x)}{r(z)} \\
& = \mathbb{E}_{x} D_{KL}(p(z|x)||r(z))
\end{align*}

\subsection{Alternative Method}

Suppose we assume that $X$ can be sampled using the process $Z_v \rightarrow X$. In practice, this would mean pre-training an unsupervised latent variable model which we will define as $p(x|z_v)$. We can think of this as an instance of the teacher-student setup, where the student is trained to mimic the teacher. Given this, we can use another approach to calculate $H(Z)$ by estimating $p(z)$ for a given $z$:
  \begin{align*}
    p(z) & = \sum \limits_x p(x, z) = \sum \limits_x p(z | x) p(x)
    = \sum \limits_x \sum_{z_v} p(z | x) p(z_v, x)
  \end{align*}
To perform variational inference, we introduce a distribution $q(z_v, x)$:
  \begin{equation}
    p(z) 
    = \sum_{z_v, x} q(z_v, x) p(z | x) \frac{p(z_v, x)}{q(z_v, x)} = \mathbb{E}_{z_v, x \sim Q} \, \, \, \, p(z | x) \frac{p(z_v, x)}{q(z_v, x)}
  \end{equation}
We can use this to lower bound $\log p(z)$:
  \begin{equation}
  \begin{aligned}
    \log p(z) &= \log \mathbb{E}_{z_v, x \sim Q} p(z | x) \frac{p(z_v, x)}{q(z_v, x)} \\
    &\ge \mathbb{E}_{x \sim Q} \log p(z | x) - \mathbb{E}_{z_v, x \sim Q} \log\frac{q(z_v, x)}{p(z_v, x)} \\
    &= \mathbb{E}_{x \sim Q} \log p(z | x) - D_{\text{KL}}(Q(z_v, x) || P(z_v, x))
  \end{aligned}
  \end{equation}
If we define $q(z_v, x)$ to use the teacher model:
$$
q(z_v, x) = q(z_v) p(x|z_v)
$$
we can simplify the KL divergence:
\begin{align*}
D_{\text{KL}}[q(z_v, x) \mid\mid p(z_v, x)]
& = \sum_{z_v, x} q(z_v) p(x \mid z_v) \log \frac{q(z_v) p(x \mid z_v)}{p(z_v) p(x \mid z_v)} \\
& = \sum_{z_v} q(z_v) \sum_{x} p(x \mid z_v) \log \frac{q(z_v)}{p(z_v)} \\
& = \sum_{z} q(z)
\log \frac{q(z_v)}{p(z_v)} \left( \sum_{x} p(x \mid z_v) \right) \\
& = D_{\text{KL}}[q(z_v) \mid\mid p(z_v)]
\end{align*}
In other words, if we have access to the "real" data generating distribution $p(x|z_v)$, we can use it during variational inference. The final equation is as follows:
\begin{equation}\label{eq:teacher-student-mi-upperbound}
\begin{aligned}
I(X, Z) \leq & - \mathbb{E}_x H[p(z|x)] \\
& -\mathbb{E}_{z, x} [\mathbb{E}_{z_v, x' \sim Q(z)} \log p(z | x') - D_{KL}(Q(z_v|x) || P(z_v))]
\end{aligned}
\end{equation}
The variational distribution $Q$ is computed for a specific $z$. Algorithmically the inference process is as follows:
\begin{enumerate}
\item Sample $z_v$, $x$ from the teacher model $p(x|z_v)$
\item Run the student encoder to get $p(z|x)$
\item Sample $z$ from $p(z|x)$
\item Run the inference network to get $q(z_v|z)$
\item Sample $z_v'$ from $q$, and re-run the teacher model $p(x'|z_v')$
\item Sample $x'$ from this distribution to compute $p(z|x')$
\item Use all the samples to compute an MI upper bound via Equation \ref{eq:teacher-student-mi-upperbound}
\item Use the upper bound as a loss function to train the inference network $q(z_v|z)$
\end{enumerate}

\subsection{Hypothesis}

The classification model we analyze is trained to maximize the IB tradeoff directly. However, the objective does not specify how this optimization will take place. We want to see whether the two training modes observed by \citet{DBLP:journals/corr/Shwartz-ZivT17} will happen for a VIB model, or whether it will train to fit and compress in an interleaved manner \citep{anonymous2018onrebut}.

Our secondary goal is to compare our two approaches to estimate mutual information of $X$ and $Z$. Both methods upper bound the true quantity $I(X, Z)$, we would like to see which methods will produce a lower result. The VIB objective has the advantage of being fast to calculate, whereas our method is optimization-based that runs over multiple iterations. However we might get a better result due to the network having access to the data generating distribution.

\section{Experiments and Results}\label{section:experiments}

We test our approach on classifying MNIST digits generated by a teacher model. The teacher is a VAE with 20 hidden states and trained for 100 epochs. The student model is a classifier trained using VIB. The hidden state $z$ is 40 dimensions and we use a Gaussian with zero mean and unit variance as $r(z)$. The decoder $p(y|z)$ is a 2-layer MLP. The encoder $p(z|x)$ can be either a 2-layered MLP or a 3-layered CNN.

To generate labeled training data, we reconstruct the training images using the teacher VAE and keep the original labels. After every epoch of training, we estimate the mutual information $I(X, T)$ from both the VIB loss function or the inference network. This network has access to the original data generating VAE.

\subsection{Zero Information Signals}

To verify our implementation, we ran training on a dataset where the images are generated independently from the labels, thereby having $I(X, Y) = 0$. We would expect to see $I(X, Z)$ and $I(Y, Z)$ to drop rapidly once the network has determined that the image contains no useful information for predicting the digit.

\begin{figure}[h]
  \centering
    \includegraphics[height=35mm]{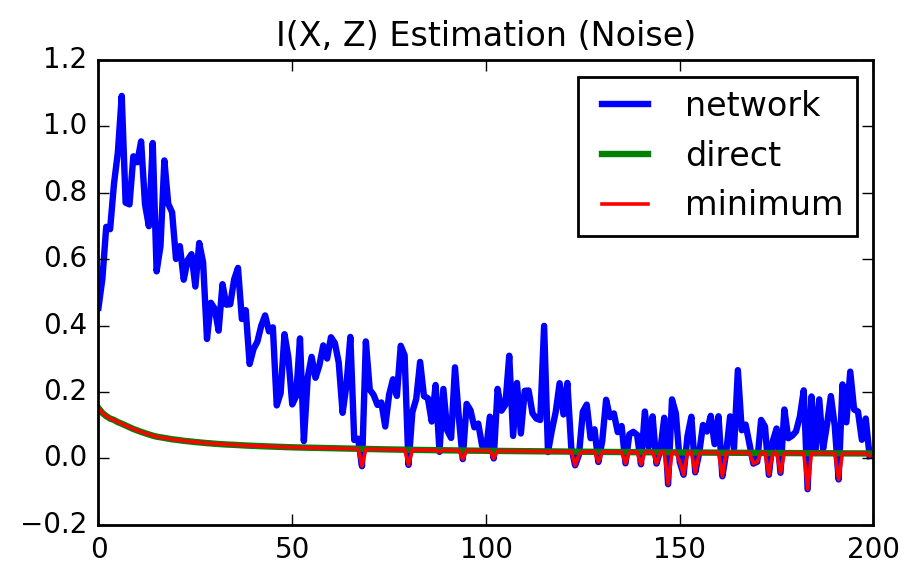}
    \includegraphics[height=35mm]{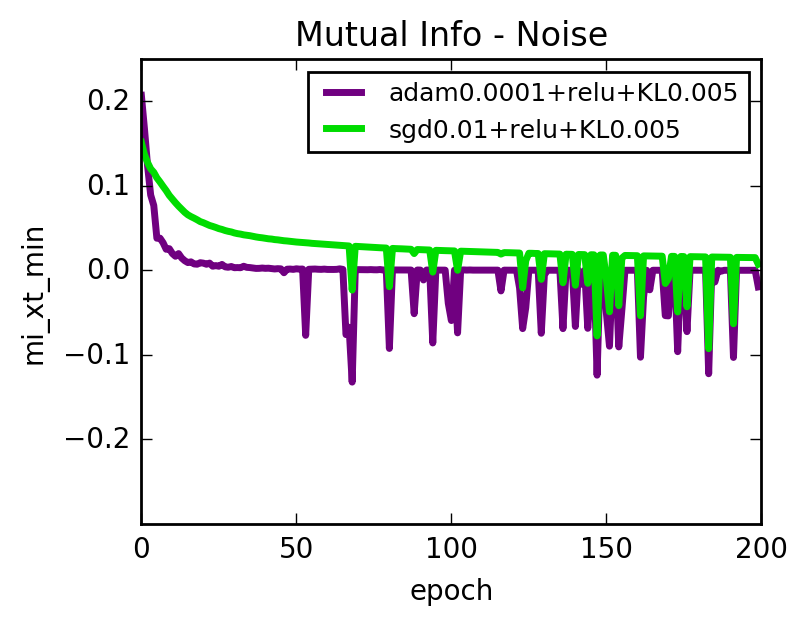}
  \caption{Results of trying to learn from images with zero information content. Both methods converge to 0, but the direct method is more stable and converges faster.}
\label{figure:3}
\end{figure}

\subsection{Comparing MI Estimators}

Our results show that the direct estimation method from \citet{DBLP:journals/corr/AlemiFD016} is better at bounding $I(X, T)$ during the later stages of training. We suspect that this is because at later stages of training, the marginal $p(z)$ is sufficiently close to the approximation $p(z)$, which produces a better estimate than doing it the roundabout way. In contrast, the inference network was able to reach a lower bound while mutual information is still rising.

\begin{figure}
  \centering
    \includegraphics[height=35mm]{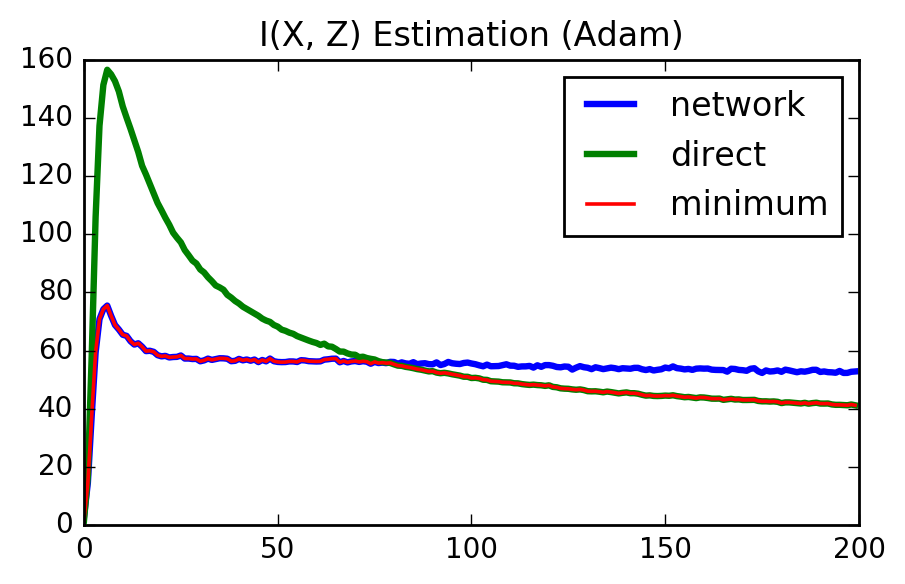}
    \includegraphics[height=35mm]{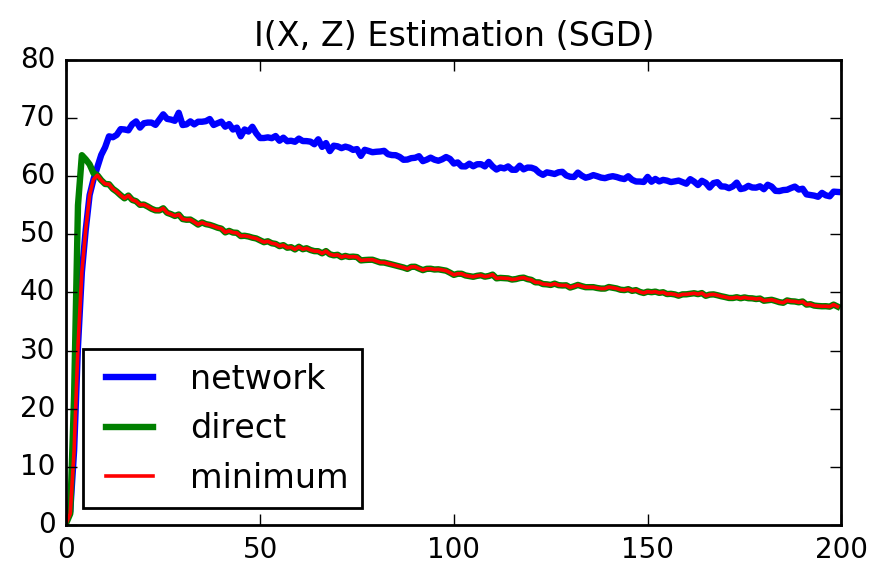}
  \caption{Comparing the two ways of measuring I(X, T). The inference network does better in the beginning but gets stuck in a local maxima, while the direct upper bound keeps going down.}
\label{figure:4}
\end{figure}

In the following sections, we will take the minimum of both estimates.

\subsection{Information Plane Dynamics}

\begin{figure}
  \centering
    \includegraphics[height=45mm]{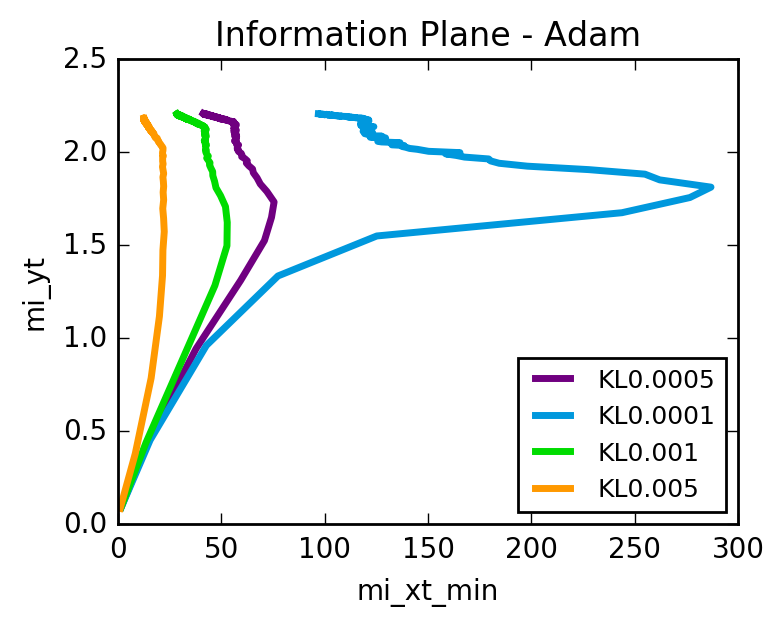}
    \includegraphics[height=45mm]{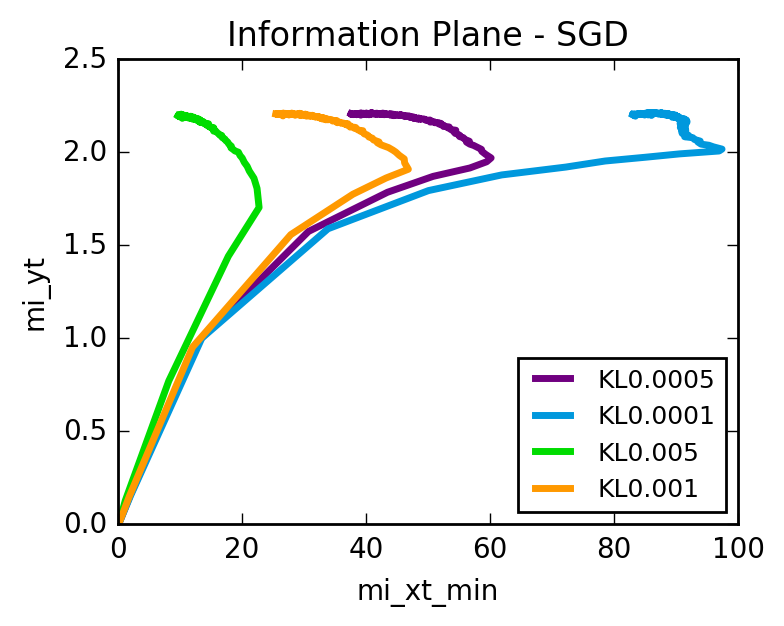}
  \caption{IP trajectory using different optimizers and different values of $\beta$. As the VIB regularization gets more aggressive, the model is more conservative about adding extra information.}
\label{figure:ip-trajectory}
\end{figure}

\begin{figure}
  \centering
    \includegraphics[height=45mm]{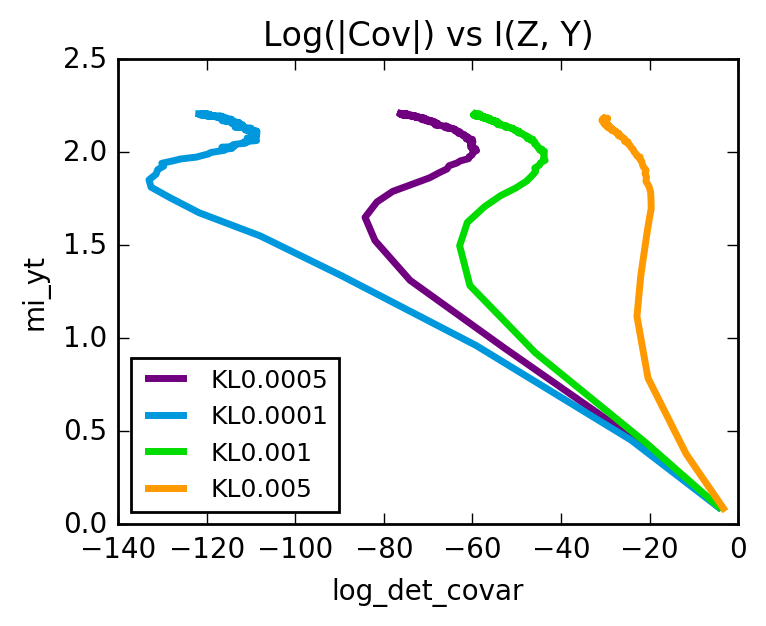}
    \includegraphics[height=45mm]{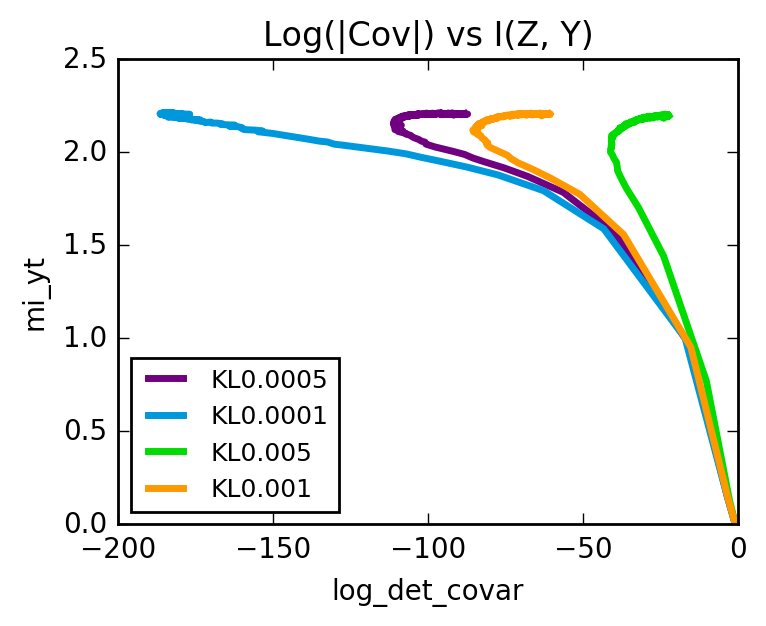}
  \caption{Plotting the average determinant of the covariance for Adam (left) and SGD (right).}
\label{figure:log-det-covar}
\end{figure}

Our experiments reveal that we were able to see an inflection point in the mutual information $I(X, T)$ as the VIB model is training (see figures \ref{figure:ip-trajectory} and \ref{figure:log-det-covar}). We see that the regularization parameter strongly affects how much the model will fit to the data before starting compression.

\subsection{Model Uncertainty}

VIB models can declare their uncertainty of an encoding given a model $p(z|x)$ by enlarging the variance. Surprisingly, during compression phase we are still able to see the variance given to real samples go down.

\begin{figure}
  \centering
    \includegraphics[height=45mm]{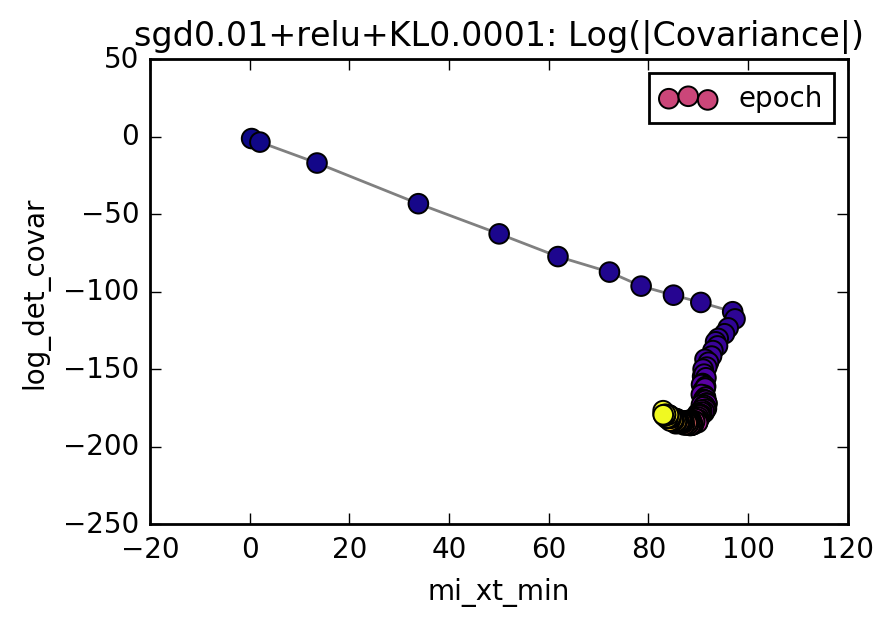}
    \includegraphics[height=45mm]{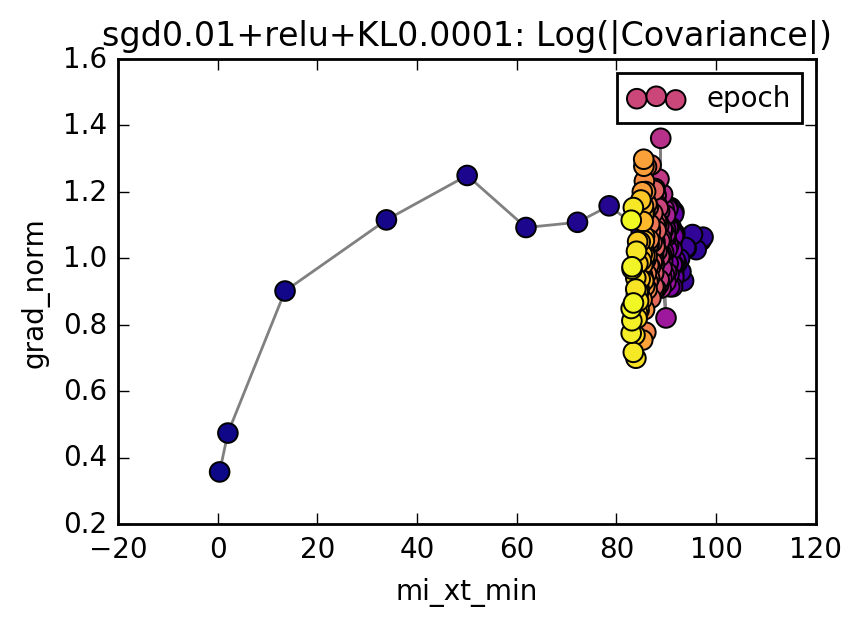}
  \caption{Covariance Determinant (left) and norm of the gradient (right) related to $(X, Z)$. Here we see that the covariance of the samples still getting tighter even as we enter the compression phase. On the right, we see that the magnitude of the gradient increases during fitting and fluctuates in the same range during compression.}
\label{figure:6}
\end{figure}

\section{Conclusion}\label{section:conclusion}

In this work we analyzed the training of a DNN on an image classification task. We confirm the existence of the two phases of SGD in the information plane for classification models whose mutual information can be calculated through variational inference. We proposed a mutual information upper bound for a teacher-student training setting and compared its performance to the bounds formulated by \citet{DBLP:journals/corr/AlemiFD016}.

Our next step is to do a baseline analysis of a linear model whose mutual information we can exactly calculate. Going in the other direction, we also wish to extend this technique to more difficult problems such as tougher image related tasks or perhaps analyzing the discriminator of a GAN. We also want to find ways to improve our mutual information lower bound estimate, and to extend this analysis to deterministic neural networks which do not train to maximize mutual information directly.

\clearpage
\begingroup
\raggedright
\bibliographystyle{plainnat}
\bibliography{main}
\endgroup

\end{document}